\documentclass[referee]{raa} 
\pdfoutput=1
\usepackage{graphicx,times}
\usepackage[a4paper, total={6in, 8in}]{geometry}
\usepackage{natbib}
\usepackage{amssymb,amsmath}
\usepackage{diagbox}
\bibpunct{(}{)}{;}{a}{}{,}
\usepackage[pagebackref=true]{hyperref}
\begin{document}
   \title{Identifying outliers in astronomical images with unsupervised machine learning}

 \volnopage{ {\bf 20XX} Vol.\ {\bf X} No. {\bf XX}, 000--000}
   \setcounter{page}{1}
   \author{Yang Han\inst{1}\thanks{First author}, 
            Zhiqiang Zou\inst{1, 2\star}, 
            Nan Li\inst{3,4}\thanks{Corresponding author, nan.li@nao.cas.cn}, 
            Yanli Chen\inst{1,2}
   }
   \institute{ School of Computer Science, Nanjing University of Posts and Telecommunications, Nanjing, Jiangsu, People’s Republic of China
        \and
             Jiangsu Key Laboratory of Big Data Security and Intelligent Processing, Nanjing, Jiangsu, People’s Republic of China\\
	\and
Key Laboratory of Optical Astronomy, National Astronomical Observatories, Chinese Academy of Sciences, Beijing, People’s Republic of China\\
\and 
University of Chinese Academy of Sciences, Beijing, People’s Republic of China\\
\vs \no
   {\small Received 20XX Month Day; accepted 20XX Month Day}
}

\abstract{Astronomical outliers, such as unusual, rare or unknown types of astronomical objects or phenomena, constantly lead to the discovery of genuinely unforeseen knowledge in astronomy. More unpredictable outliers will be uncovered in principle with the increment of the coverage and quality of upcoming survey data. However, it is a severe challenge to mine rare and unexpected targets from enormous data with human inspection due to a significant workload. Supervised learning is also unsuitable for this purpose since designing proper training sets for unanticipated signals is unworkable. Motivated by these challenges, we adopt unsupervised machine learning approaches to identify outliers in the data of galaxy images to explore the paths for detecting astronomical outliers. For comparison, we construct three methods, which are built upon the k-nearest neighbors (KNN), Convolutional Auto-Encoder (CAE) $+$ KNN, and CAE $+$ KNN $+$ Attention Mechanism (attCAE$\_$KNN) separately. Testing sets are created based on the Galaxy Zoo image data published online to evaluate the performance of the above methods. Results show that attCAE$\_$KNN achieves the best recall (78$\%$), which is 53$\%$ higher than the classical KNN method and 22$\%$ higher than CAE$+$KNN. The efficiency of attCAE$\_$KNN (10 minutes) is also superior to KNN (4 hours) and equal to CAE$+$KNN (10 minutes) for accomplishing the same task. Thus, we believe it is feasible to detect astronomical outliers in the data of galaxy images in an unsupervised manner. Next, we will apply attCAE$\_$KNN to available survey datasets to assess its applicability and reliability.
}

   \authorrunning{Yang Han et al.}            
   \titlerunning{Identifying outliers in astronomical images}  
   \maketitle

%
\section{INTRODUCTION}           
\label{sect:intro}
Astronomy is stepping into the big data era with the upcoming large-scale surveys \citep{2021Personalisedactive}, e.g., Euclid \footnote{\url{https://www.euclid-ec.org/}}, LSST \footnote{\url{https://www.lsst.org}} and CSST \footnote{\url{http://www.bao.ac.cn/csst/}}. Mining knowledge from enormous astronomical datasets has become critical for astrophysical and cosmological investigations . Typically, data mining in astronomy includes object classification, dependency detection, class description, and anomalies/outlier detection. The first three categories of tasks are problem-driven, i.e., once the goals are well-defined, the tasks can be handled in a supervised manner by involving well-designed training sets. These tasks help improve the accuracy and precision of the models for describing mainstream objects, and relevant approaches are relatively mature and widely applied in astronomy \citep{lukic2019, yy26+zhu2019galaxy, yy16+cheng2020identifying, Gupta2022, Chen_2022, zhang2022classifying}. On the other hand, astronomical anomalies/outliers constantly lead to unforeseen knowledge in astronomy, which may trigger revolutionary discoveries. Expectedly, more unpredictable outliers should be uncovered in principle with the increment of the coverage and quality of upcoming survey data. Therefore, developing approaches for outlier detection are as important as those for the first three tasks \citep{2020TransformationBased, 2021Activeanomaly, 2020Unsupervisedmachine}.

Outliers are defined in various papers \citep{Hawkins1980, Beckman&Cook1983, Barnett&Lewis1987, Tabachnick&Fidell1996}, generally, it is described as: an outlier is an observation that deviates significantly from primary observations so that it aroused suspicions that a different mechanism generates it. In daily life, outlier detection has numerous applications, including credit card fraud detection, the discovery of criminal activities in E-commerce, video surveillance, pharmaceutical research, weather prediction and the analysis of performance statistics of professional athletes. Most of them are relevant to troubles. Nevertheless, the detection of astronomical outliers always leads to the discovery of surprising unforeseen facts and expands the boundaries of human knowledge of the universe \citep{yy1+pruzhinskaya2019anomaly, 2019DetectingOutliers}. Hence, it is necessary to develop efficient and automated approaches for detecting astronomical outliers and understand their feasibility and reliability thoroughly, particularly in the era of big data \citep{Outlierdetection2004, 2020MNRASDetectingoutlier}.

Early outlier detection methods  \citep{yy3+edgeworth1888some, yy4+zhang2004outlier, yy5+dutta2007distributed,yy6+solarz2017automated,yy7+giles2019systematic, yy8+fustes2013approach, yy9+baron2017weirdest} are generally based on traditional unsupervised learning algorithms. For instance, \cite{yy7+giles2019systematic} employed a variant of the DBSCAN clustering algorithm to detect outliers in derived light curve features. \cite{yy9+baron2017weirdest} extracted the feature of galaxy spectra manually firstly and then adopted an unsupervised Random Forest to detect the most outlying galaxy spectra within the Sloan Digital Sky Survey \footnote{\url{https://www.sdss.org/}} (SDSS). Moreover, as a well-known clustering algorithm, k-Nearest Neighbor \citep{yy10+dasarathy1991nearest} becomes popular for detecting outliers since it operates without assumption about the data distribution. However, these traditional methods become unsuitable when the volume and quality of astronomical images increase greatly. One reason is that the feature extraction routines in traditional methods are too coarse and inflexible to retain details and untypical features of the high-quality astronomical images; another reason is that the efficiency of CPU-based traditional methods is too slow to handle the tremendous volume of future survey data.

Recently, beyond traditional machine learning, deep learning is utilized to construct programs for detecting outliers \citep{yy11+chalapathy2019deep, yy12+nadeem2016semi, yy13+hendrycks2018deep, 2021AnomalyDetection}, such as Auto-Encoder \citep[AE ][]{yy14+vincent2010stacked} and Convolutional Auto-Encoder \citep[CAE][]{yy15+masci2011stacked, 2020AnomalyDetection}. AE and CAE represent input images with a feature vector which can be used to reconstruct the input images with the most likelihoods. This feature extraction procedure is automated and speedy. Besides,  Bayesian Gaussian Mixture is utilized to implement the clustering process and then identify the galaxy images' outlier according to the distribution of the feature vectors in latent space \citep{2021Beyond}.
Combining the above two modules, one can classify galaxy images without labels \citep{yy16+cheng2020identifying}, as well as to detect outliers. However, the performance of such unsupervised approaches based on deep learning is above 10\%\textasciitilde20\% worse than that of supervised approaches due to noisy data \citep{yy17+zhou2017anomaly}. 

In this work,  we adopt the attention mechanism \citep{yy18+vaswani2017attention} to further improve the performance of the unsupervised methods as it can make the CAE pay more attention to the critical features and suppress background noise. To understand the differences from traditional outlier detection methods to state-of-art attention-improved ones systematically, we construct three programs, which are built upon the KNN, CAE $+$ KNN and CAE $+$ KNN $+$ Attention mechanism, separately. We organize two types of datasets based on the galaxy images data published by the Galaxy Zoo Challenge Project on Kaggle \footnote{\url{https://www.kaggle.com/c/galaxy-zoo-the-galaxy-challenge}} to evaluate the performance of various approaches in different cases. The first datasets of galaxy images for testing the above approaches include inliers containing a single type of galaxy morphology plus outliers containing a single type of galaxy morphology; the second dataset is similar to the first ones but with multiple types of galaxy morphology in the outliers. After conducting extensive experiments, we find that CAE boosts the clustering process significantly and improves the accuracy of detecting outliers; the attention mechanism increases the accuracy further since it guarantees CAE to extract valuable features only, avoiding noise. It is the first time involving the attention mechanism in the outlier detection of astronomical images, which is worth being included in the program for similar purposes in the future. For the convenience of other researchers, we published the code and data used in this project onine \footnote{\url{https://github.com/hanlaomao/hanlaomao.git}}.

This paper is structured as follows. We introduce the datasets used in this work in Sect.~\ref{sect:data}. Sect.~\ref{sect:method} describes the methods we constructed. Details about the experiments, including data processing and the implementation of outliers detection with the above approaches, are shown in Sect.~\ref{sect:experiment}. We summarize and analyze the results in Sect.~\ref{sect:results}. Finally, the discussion and conclusions are delivered in Sect.~\ref{sect:discussion}.

\section{DATA}
\label{sect:data}
The galaxy morphology data used in this study is collected from the Galaxy Zoo project \citep{yy2+willett2013galaxy, yy19+ventura2011hot}. In this section, we first introduce the origin and composition of the dataset and then present the filtering methods and how to divide the original data in Sect. \ref{sect:2.1}. Sect.~\ref{sect:2.2} describes how to construct experimental data subsets to evaluate the performance of outlier detection with the approaches mentioned in Sect.~\ref{sect:method}.

\subsection{The Galaxy Zoo Dataset}
\label{sect:2.1}
The SDSS captured around one million galaxy images. To classify these galaxies morphologically,
the Galaxy Zoo Project was launched \citep{lintott2008galaxy}, a crowd-sourced astronomy project inviting volunteers to assist in the morphological classification of large numbers of galaxies. The dataset we adopted is one of the legacies of the galaxy zoo project, and it is publicly available online with the Galaxy-zoo Data Challenge Project on Kaggle\footnote{\url{https://www.kaggle.com/c/galaxy-zoo-the-galaxy-challenge/data}}. 

The dataset provides 28793 galaxy morphology images with middle filters available in SDSS (g, r, and i) and a truth table including 37 parameters for describing the morphology of each galaxy. The 37 parameters are between 0 and 1 to represent the probability distribution of galaxy morphology in 11 tasks and 37 responses \citep{yy2+willett2013galaxy}. Higher response values indicate that more people recognize the corresponding features in the images of given galaxies. The catalog is further debiased to match a more consistent question tree of galaxy morphology classification \citep{hart2016galaxy}.

To make the problem of outlier detection representative, we reorganize 28793 images into five categories: Completely round smooth, In-between smooth, Cigar-shaped smooth,Edge-on and Spiral  according to the 37 parameters in the truth table. The filtering method refers to the threshold discrimination criteria in \cite{yy26+zhu2019galaxy}. For instance, when selecting the Completely round smooth, values are chosen as follows: $f_{smooth}$ more than 0.469, $f_{complete,round}$ more than 0.50, as shown in Table~\ref{tab1}. The testing sets are constructed by choosing images from the above categories, and details are presented in Sect.~\ref{sect:2.2}.

\begin{table}[]
\caption{The five galaxy morphology categories from 0\textasciitilde4 with 28793 samples. The first column is the category id, the second column is the name of category, the third column is the thresholds corresponding to each category, and the last column is the number of galaxy images in each category.}\label{tab1}
    \centering
\begin{tabular}{cclc}
\hline
Category    &Category name              & Thresholds                & Number \\ \hline
    0            &Completely round smooth & $f_{smooth} \geq 0.469        $   & 8436   \\ \hline
                 &                       & $f_{completely round} \geq 0.50 $ &        \\
    1            &In-between smooth       & $f_{smooth} \geq 0.469        $   & 8069   \\
                 &                       & $f_{in-between} \geq 0.50     $   &        \\ \hline
    2            &Cigar-shaped smooth     & $f_{smooth} \geq 0.469        $   & 579    \\
                 &                       & $f_{cigar-shaped} \geq 0.50   $   &        \\ \hline
    3            &Edge-on                 & $f_{features/disk} \geq 0.430 $   & 3903   \\
                 &                       & $f_{edge-on,yes} \geq 0.602   $   &        \\ \hline
    4            &Spiral                  & $f_{edge-on,no} \geq 0.715    $   & 7806   \\
                 &                       & $f_{Spiral,yes} \geq 0.619    $   &        \\ \hline
Total            &                          &                              & 28793   \\
                        \hline

\end{tabular}
\end{table}

\subsection{Experimental Data Subsets}
\label{sect:2.2}
To mimic different cases of outlier detection, we construct two sorts of experimental data subsets by selecting images from the categories of galaxy images described in Sect.~\ref{sect:2.1}. One group of testing sets includes inliers containing a single type of galaxy morphology plus outliers containing a single type of galaxy morphology; the other group of testing sets is similar to the first ones for inliers but containing multiple types of galaxy morphology in the outliers.

Implicitly, the first group contains four data subsets, the inliers are all Completely round smooth galaxies, and the outliers are selected from other categories of galaxies separately. The fraction of outliers is $10\%$ in each subset. The second group contains one data subset, the inliers are also Completely round smooth galaxies, but the outliers consist of galaxy images from other categories of galaxies. The total fraction of outliers is $10\%$ as well, and the four types of galaxy images are equally constituted in the outliers.

Table~\ref{tab2} shows an overview of the above five testing sets, and columns denote the structure of each testing set. For instance, the first testing set (subsect1) consists of Completely round smooth galaxies (category 0) as inliers and Cigar-shaped smooth galaxies (category 2) as outliers. There are 16000 inliers and 1778 outliers. Note that when lacking galaxy images of some categories, we expand the insufficient number of galaxy images by using data augmentation (see Sect.~\ref{sect:4.1}).

\begin{table}[]
\caption{Number of samples in five different experimental data subsets. The first column is the category id and the last five columns are five date subsets.  For example, the second column represents the first experimental date subset, it just contains 16000 samples of category 0 as inliers and 1778 samples of category 2 as outliers. The last column represents the fifth data subset which contains 16000 samples of category 0 as inliers and contains total 1778 samples from catefory 1\textasciitilde4.}\label{tab2}
    \centering
\begin{tabular}{cccccc}
\hline
Category    &Subset1                &Subset2            &Subset3                &Subset4            &Subset5\\ \hline
0           &16000                  &16000              &16000                  &16000              &16000\\ \hline
1           & 0                     &0                  &1778                   &0                  &445\\ \hline
2           &1778                   &0                  &0                      &0                  &445 \\ \hline
3           &0                      &1778               &0                      &0                  &444\\ \hline
4           &0                      &0                  &0                      &1778               &444\\ \hline
Total       &17778                  &17778              &17778                  &17778              &17778\\ \hline

\end{tabular}
\end{table}

\section{METHODOLOGY}
\label{sect:method}
For comparing the traditional methods and our state of art method, we build three approaches for outlier detection. The simplest one is based on KNN only, a classic clustering algorithm grounded on distance metrics. The second one involves CAE for feature extraction but still utilizes KNN for the clustering procedure. At last, we employ the attention mechanism to improve the stability of the feature extraction with CAE. The following subsections demonstrate details of the construction of these approaches.
\subsection{The KNN-based Approach}
\label{sect:3.1}
\begin{figure}
\centering
\includegraphics[width=\hsize]{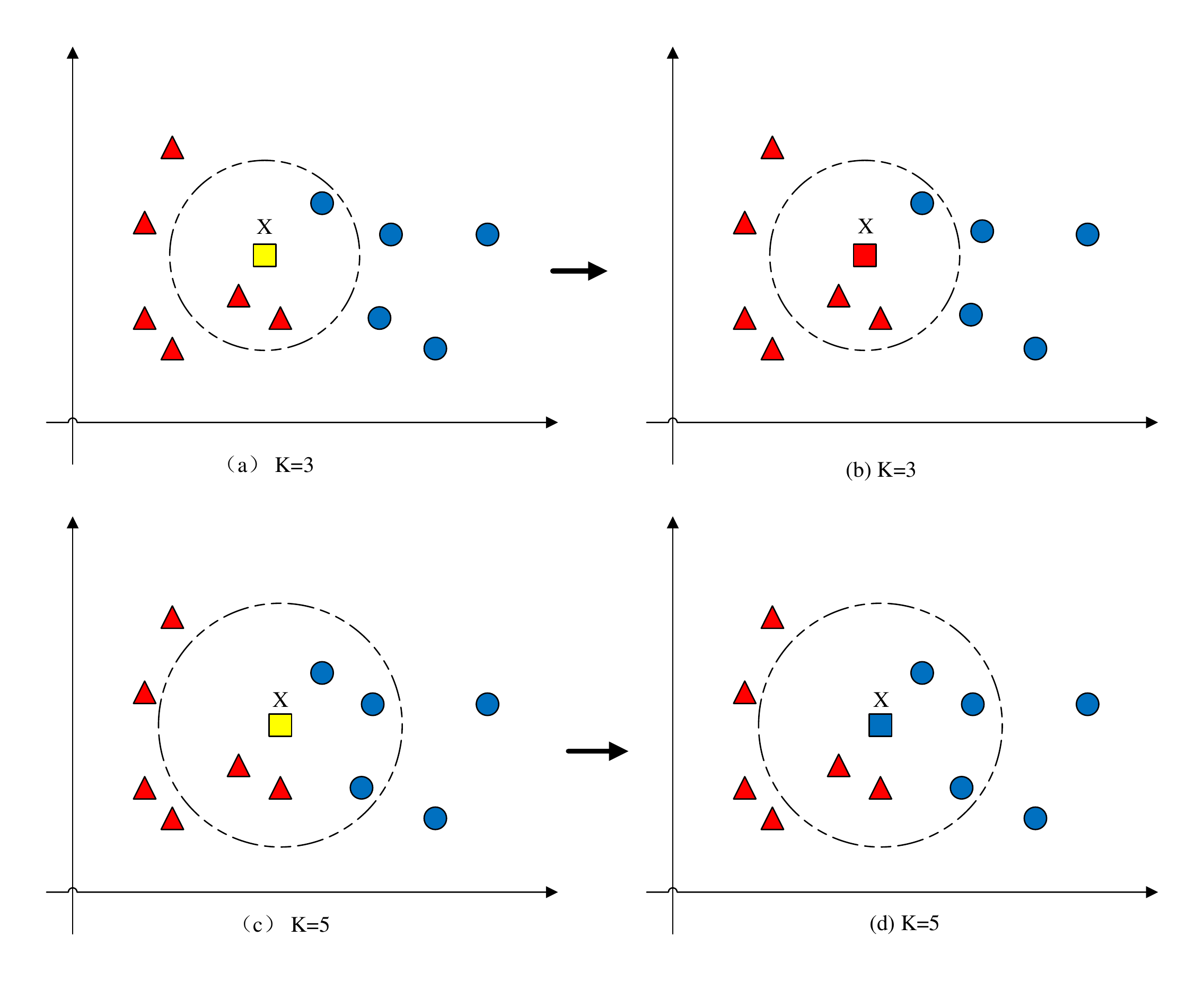}
\caption{The classifying results based on classical KNN.  Panel (a) and panel (b) present the procedure of an element with yellow rectangle is classified to the category with red triangle when K=3. Panel (c) and panel (d) describe the procedure of data X with yellow rectangle is classified to the category with blue circle when K=5.}
\label{fknnflow}
\end{figure}

The KNN algorithm is one of the non-parametric classifying algorithms \citep{yy10+dasarathy1991nearest}, whose core idea is to assume that data X has K nearest neighbors in the feature space. If most K neighbors belong to a certain category, the X could also be determined to belong to this category. As shown in Fig.~\ref{fknnflow} (a), the yellow rectangle is the data X needs to be predicted. Assuming K=3, as shown in Fig.~\ref{fknnflow} (b), then the KNN algorithm will find the three neighbors closest to X (here enclosed in a circle) and select a category with the most elements. For example, in Fig.~\ref{fknnflow} (b), there are more elements described by red triangles, so the X is classified to the category containing elements described by red triangles. As shown in Fig.~\ref{fknnflow} (c) and Fig.~\ref{fknnflow} (d), when K=5, the X is classified to the category containing elements described by blue circles. \cite{yy20+hu2019introductory} used KNN-based algorithms to perform classification experiments on a variety of datasets and achieved good results without any assumptions about the data. However, the KNN-based algorithm would cost considerable computing time due to the high data dimension in the case of astronomical images as input data.

\subsection{The CAE-KNN-based Approach}
\label{sect:3.2}
\begin{figure}
\centering
\includegraphics[width=\hsize]{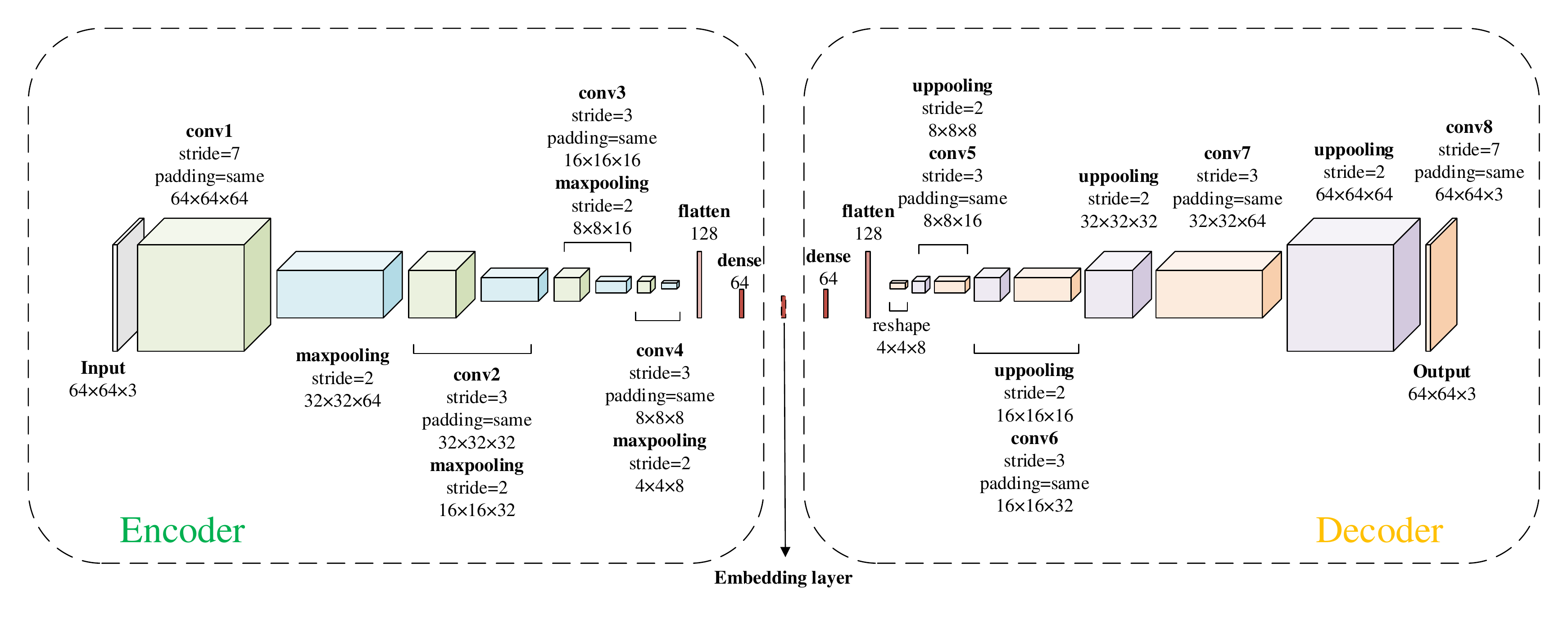}
\caption{The architecture and components of CAE. CAE consists of two components, the encoder and the decoder. Green cuboids in encoder and orange cuboids in decoder denote convolution layer; blue cuboids indicate maxpooling layer; purple cuboids present uppooling layer.}
\label{fcaearc}
\end{figure}

    CAE \citep{yy15+masci2011stacked} is an optimized AE by adopting a convolution operation, which could extract principal features of astronomical image with high dimension. CAE$\_$KNN makes full use of the CAE advantage in reducing the dimension to improve above KNN-based algorithm. We first present the architecture and components of CAE as shown in Fig.~\ref{fcaearc}, and then describe the joint of CAE and KNN.

CAE consists of two components: the encoder and the decoder. The first component is the encoder, which is responsible for extracting the representative features from input images. For an input image $x$, the $j^{th}$ representative feature map $h^{j}$ is expressed as Eq.~\ref{eq1}.
\begin{equation}
    \label{eq1}
   h^{j}=f(x*W^{j}+b^{j}) ,
\end{equation}
where $W^{j}$ is the $j^{th}$ filter, $*$ denotes the convolution operation, $b^{j}$ is the corresponding bias of the feature map and \emph{f} is an activation function. The activation function $f(z)$, where the input denotes by $z$  used in the convolutional layers, is the Rectified Linear Unit (ReLu) \citep{yy21+bengio2007scaling}, as described in Eq.~\ref{eq2}.
\begin{equation}
    \label{eq2}
   f(z)=\begin{cases}
0 \; \; \; \; if\; z\; < \; 0\\
z \; \; \; \; if\; z\; \geq \; 0  .
    \end{cases}
\end{equation}
The encoder in this study is built with four convolutional layers (filter size: 64, 32, 16, and 8) and three dense layers (unit size: 128, 64, 32). A pooling layer follows each convolutional layer with 2 by 2 pixels. The pooling layer is also considered a down-sampling layer, aiming to reduce the volume of parameters involved in the encoder.

The second component of the CAE is the decoder, and its function is to reconstruct the input image according to the extracted feature map obtained by the encoder. The decoder structure is symmetrical with the structure of the encoder. In other words, its structure is just the opposite of the encoder structure. As for the detail of reconstructing procedure, please refer to  \citep{yy15+masci2011stacked, yy16+cheng2020identifying}. The decoder has three dense layers (unit size: 32, 64, and 128), four convolutional layers (filter size: 8, 16, 32 and 64) using the ReLu activation function  \citep{yy21+bengio2007scaling}, and an extra convolutional layer (filter size: 3) using the softmax \citep{yy22+ren2017robust} function as the output of the decoder. Except for the last output layer, there is an upsampling layer behind each convolutional layer, whose function is to gradually restore the feature maps to the same shape as input images. The layer between the encoder and the decoder is the embedding layer used to reconstruct the input galaxy images.

The loss function  $L$ between the two components is given by Eq.~\ref{eq3} \citep{yy16+cheng2020identifying}.

\begin{equation}
    \label{eq3}
   L=-\frac{1}{N}\left [ t^{n}log{y^{n}}+ \left ( 1-t^{n} \right )log\left ( 1-y^{n} \right )\right ]\,\,,
\end{equation}
where $N$ is the number of samples, $t^{n}$ is the target data, and $y^{n}$ is the reconstructed data. The goal of CAE is to minimize the reconstruction error by using loss function $L$.

As so far, we could get the low dimension features from galaxy images by using the embedding layer vectors in CAE. And then, these features are fed into the KNN algorithm avoiding the time-consuming problem of KNN outlier detection. However, the CAE$\_$KNN has the disadvantage of instability since the background noise of the galaxy image sometimes influences the stability of the outlier detection.
\subsection{The Attention-CAE-KNN-based Approach}
\label{sect:3.3}
\begin{figure}
\centering
\includegraphics[width=\hsize]{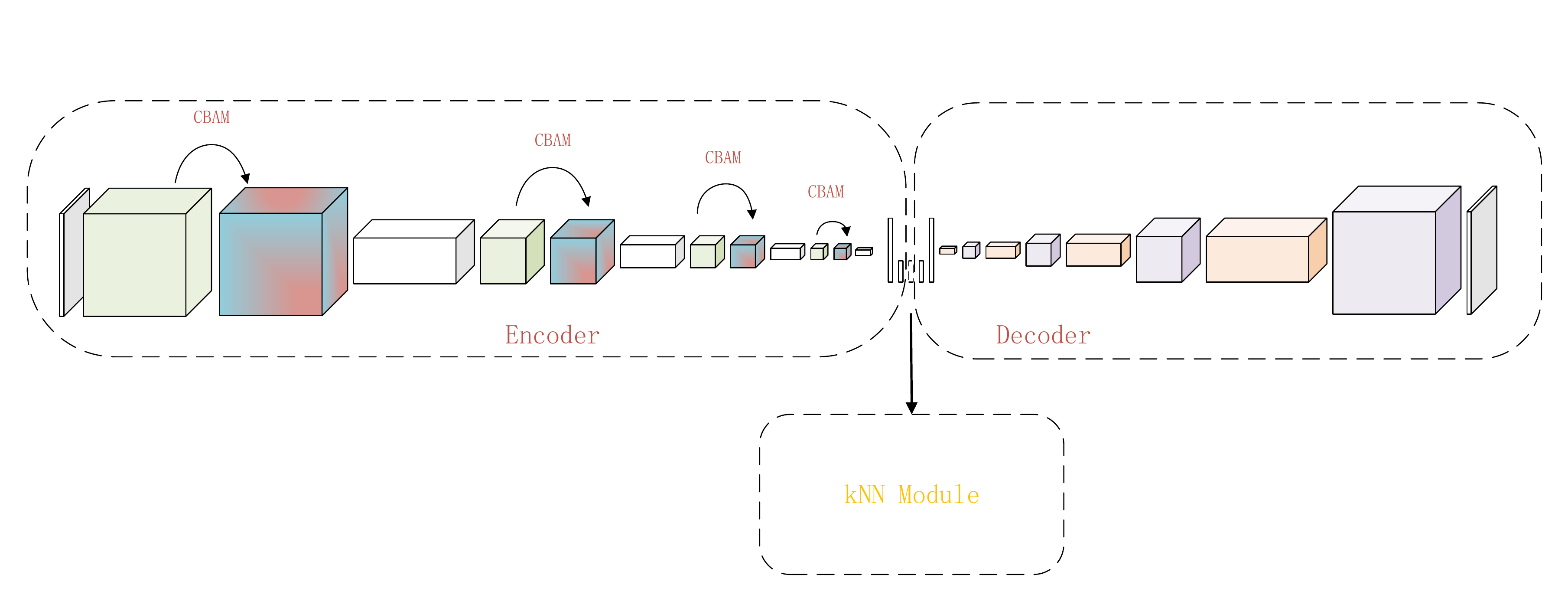}
\caption{The architecture of the attCAE$\_$KNN, including encoder, decoder and KNN module, where the CBAM attention strategy is added to the encoder.}
\label{fattarc}
\end{figure}

To increase the stability of the CAE$\_$KNN, we propose a novel algorithm, namely attCAE$\_$KNN, which is the first time to explore the attention strategy to CAE. Attention strategy \citep{yy23+xu2015show, yy24+gregor2015draw} makes the attCAE$\_$KNN focus on 'what' is meaningful for given astronomical images so that attCAE$\_$KNN could ignore the background noise. We build attCAE$\_$KNN by adopting a convolutional block attention module \citep[CBAM][]{yy25+liu2019rgb}. Its architecture is shown in Fig.~\ref{fattarc}, including encoder, decoder, and KNN module. The decoder and KNN module have been described in Sect.~\ref{sect:3.1} and Sect. \ref{sect:3.2}. Next, we focus on the improved encoder. 

\begin{figure}
\centering
\includegraphics[width=\hsize]{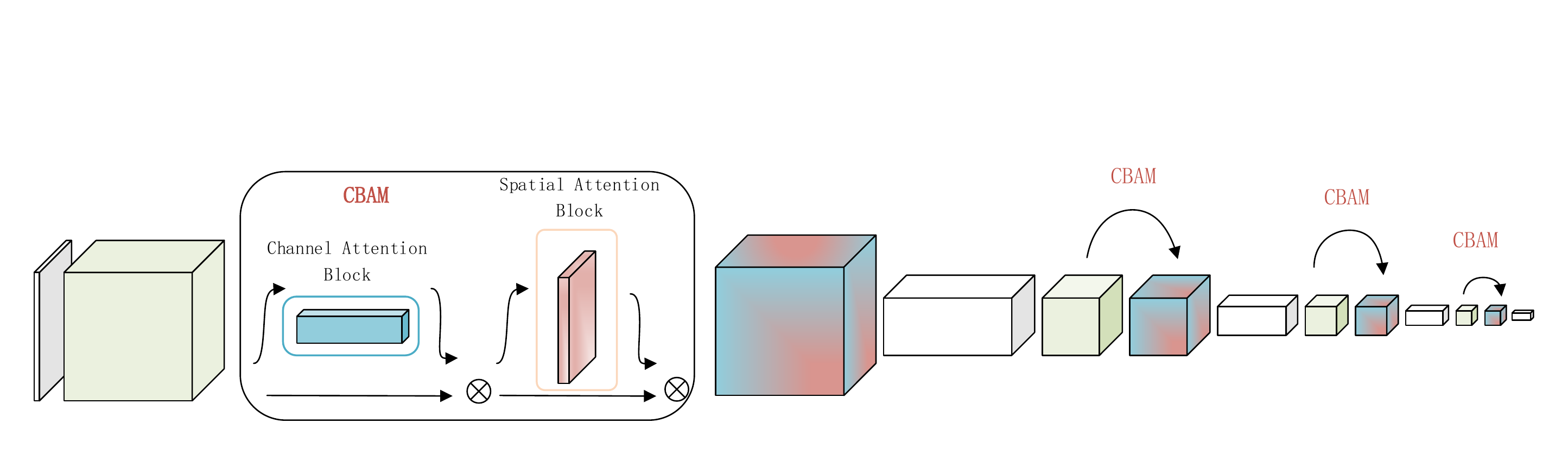}
\caption{The encoder of the attCAE$\_$KNN, which consists of the channel attention block, the spatial attention block and other CAE blocks.}
\label{fattencoder}
\end{figure}

The first part is the encoder that consists of the channel attention block and the spatial attention block \citep{yy25+liu2019rgb}, which differs from the classical encoder in inserting CBAM, as shown in Fig.~\ref{fattencoder}. These two blocks can extract the meaningful features of astronomical images along the two dimensions of the channel axis and the spatial axis. 
The second part is the decoder, in which the CBAM is not inserted after the convolutional layer. This is because through the analysis of experimental results, adding the CBAM after the convolutional layer of the decoder hardly improves the experimental results. To reduce model complexity and decrease model training time, we only add the CBAM after the convolutional layer in the encoder of the CAE. The last part is the KNN module, whose input data is the latent features from the embedding layer of attCAE.

\section{EXPERIMENT} 
\label{sect:experiment}

We present the details of experiments with the data and methods described in Sect.~\ref{sect:data} and Sect.~\ref{sect:method} here. It includes data processing, parameters of the machine learning models, evaluation metrics and experimental environments.

\subsection{Data Pre-processing}
\label{sect:4.1}
\begin{figure}
\centering
\includegraphics[width=\hsize]{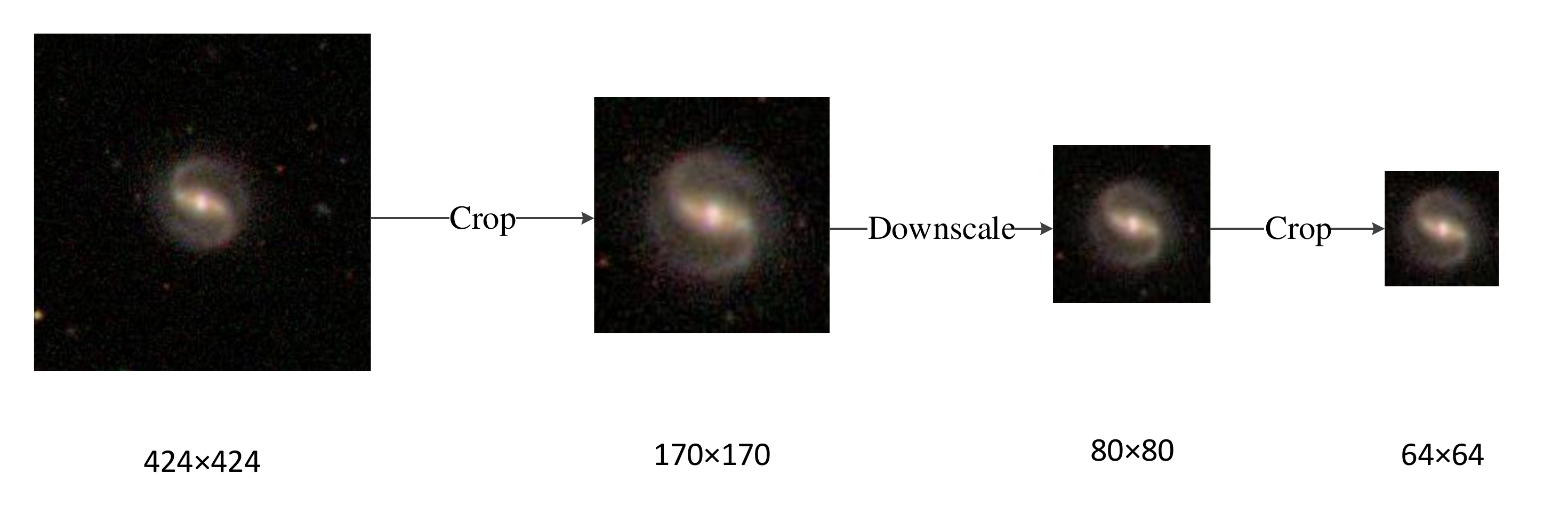}
\caption{The procedure of data preprocessing  on the original galaxy image.}
\label{fdataaug}
\end{figure}

\begin{figure}
\centering
\includegraphics[width=\hsize]{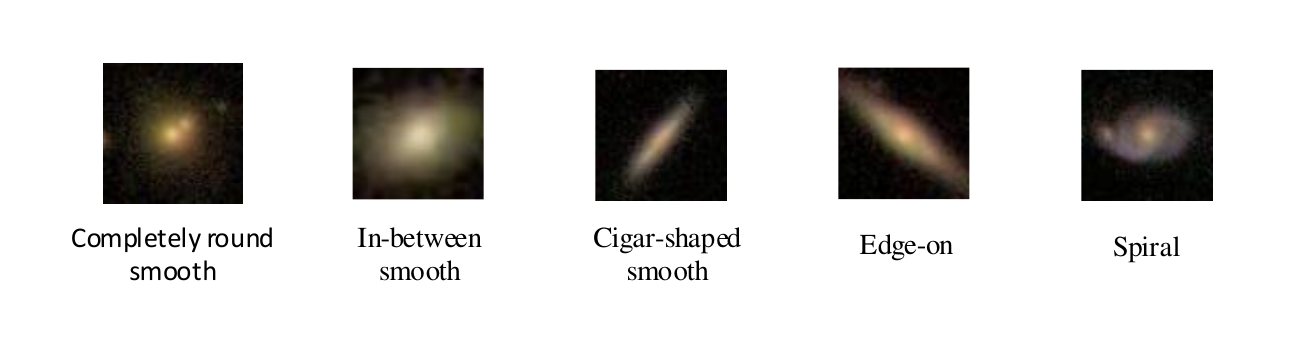}
\caption{Five representative examples from five categories.}
\label{ffiveexamples}
\end{figure}

As is shown in Sect.~\ref{sect:data}, we obtain 28793 RGB color images with a size of $424\times424\times3$ pixels.  Considering the valuable features of these images are concentrated at the central part, we conduct some pre-processing operations (see Fig.~\ref{fdataaug}). The first step is to crop the images with a box of $170\times170$ pixels in all channels. The second step is to downscale images from $170\times170\times3$ pixels to $80\times80\times3$ pixels. The last step is to crop images from $80\times80\times3$ pixels to $64\times64\times3$ pixels further. The detailed operations refer to the process in \citep{yy26+zhu2019galaxy}. Five processed examples from five categories described in Sect.~\ref{sect:data} are displayed in Fig.~\ref{ffiveexamples}.

The number of images in some categories is too small to be outliers for supporting machine learning algorithms for outlier detection, for instance, there are only 579 Cigar-shaped smooth galaxies. Thus, we make data augmentation by rotating these images randomly and finally obtain 17778 images in five data subsets, where each data subset consists of 16000 inliers and 1778 outliers (see Table~\ref{tab2}).

\subsection{Training And Clustering}
\label{sect:4.2}

\begin{figure}
\centering
\includegraphics[width=\hsize]{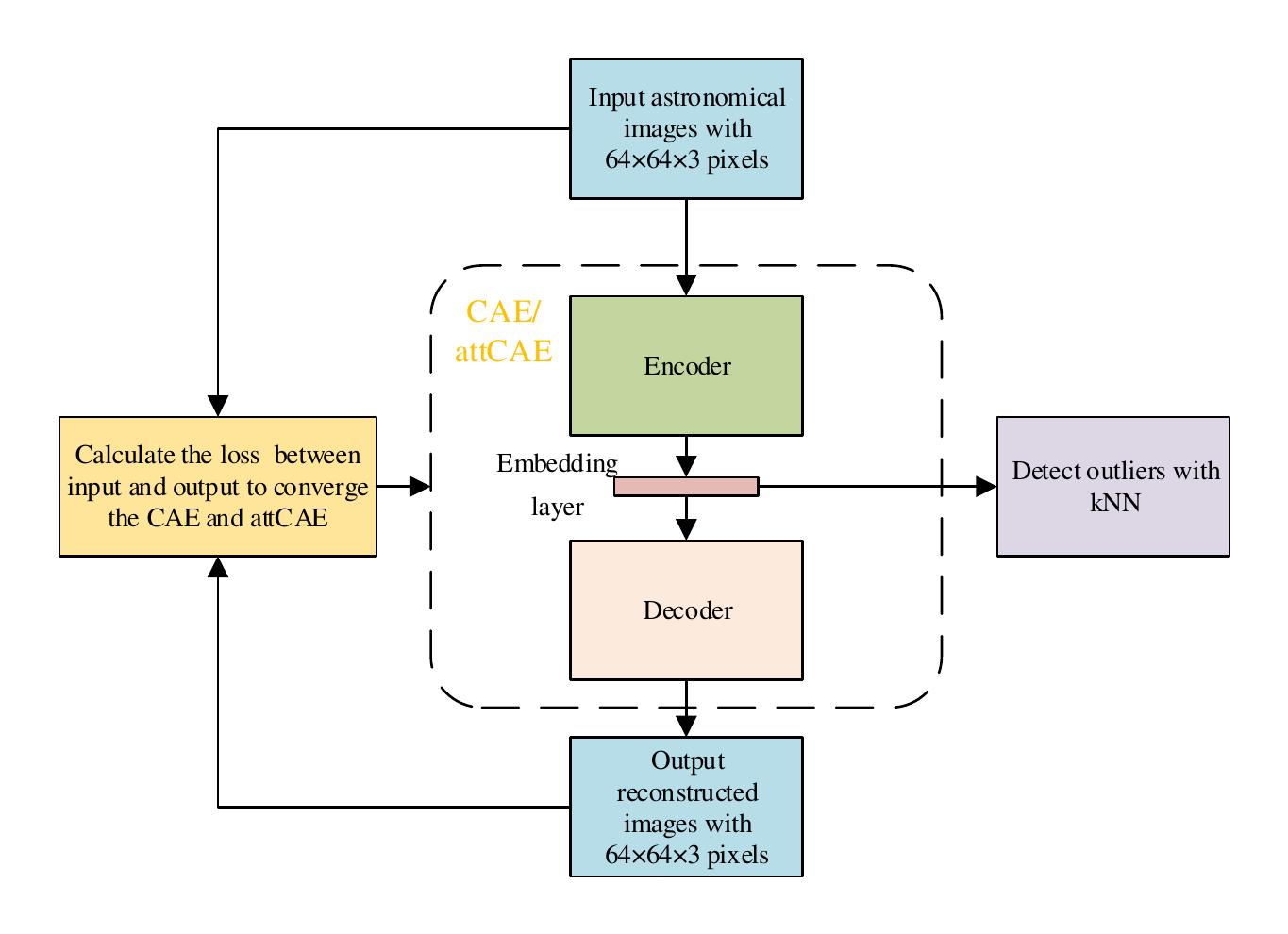}
\caption{The flow chart of the attCAE$\_$KNN for detecting outliers in astronomical images.}
\label{fflowchartall}
\end{figure}

We apply the three methods (KNN, CAE$\_$KNN, and attCAE$\_$KNN) to the data subsets separately. The training process consists of auto-encoder training and KNN training. The former is for extracting the representative features of the astronomical images, while the latter is for detecting outliers. The flow chart of the attCAE$\_$KNN for detecting outliers in astronomical images is shown in Fig.~\ref{fflowchartall}.

The training process in this paper is entirely different from the training process in the context of supervised learning. We train CAE to extract features by comparing the input images and generated images, so no labels are included in the whole process. To avoid overfitting, we divide each dataset shown in Table.~\ref{tab2} into training sets and testing sets with a ratio of 7:3, and the images in the training set and test set are randomly selected from the whole set with 17778 images. Considering that the number of outliers always accounts for a small part of the total dataset, we set the proportion of outlier data to account for 10\% of the whole dataset for detecting outliers. For example, the number of outliers in the test set is 533, which can be calculated by 17778×0.3×0.1. 

\begin{figure}
\centering
\includegraphics[width=\hsize]{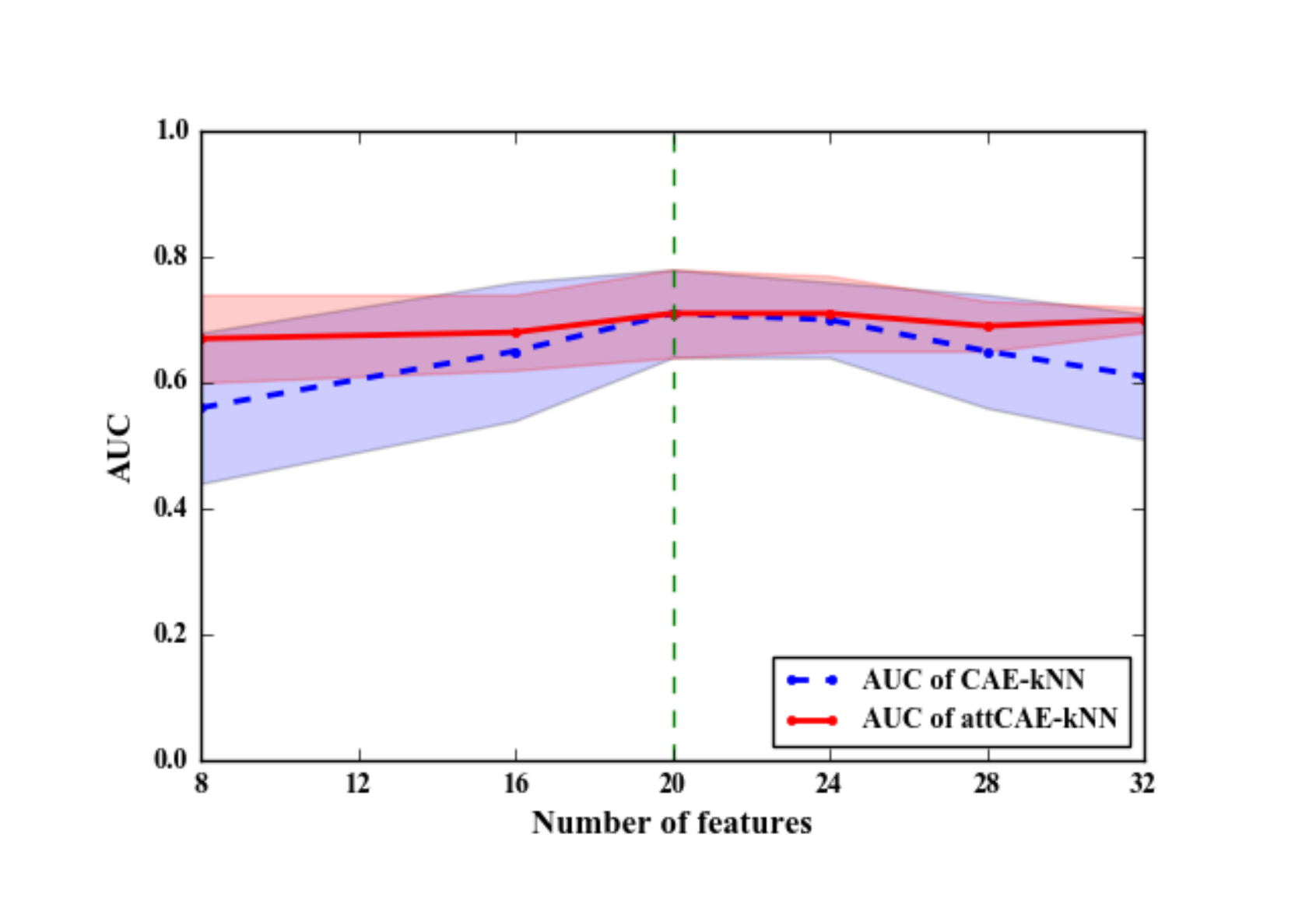}
\caption{Effect on AUC mean values of various feature numbers in embedding layer.}
\label{fauceffects}
\end{figure}

During the training procedure of CAE, parameters of the embedded layer need to be optimized in a data-driven manner. We use area under the receiver operating characteristic  curve \citep[AUC][]{yy29+bradley1997use, yy30+fawcett2006introduction}) as the criteria. The receiver operating characteristic \citep[ROC][]{yy30+fawcett2006introduction, yy16+cheng2020identifying} can be drawn with false-positive rates ($FPR$) and true-positive rates ($TPR$), which are given by Eq. (\ref{eq4}) and Eq. (\ref{eq5}), 
\begin{equation}
    \label{eq4}
    FPR=\frac{FP}{FP+TN} ,
\end{equation}
\begin{equation}
    \label{eq5}
   TPR=\frac{TP}{TP+FN} ,
\end{equation}
where $TP$ means true positive, $TN$ means true negative, $FP$ denotes false positive and $FN$ means false negative, respectively.
We then repeat the outlier detection process and compare the AUC of each classification to find the most optimal number of extracted features within the embedding layer in the CAE. In Fig.~\ref{fauceffects}, the blue dashed line shows the mean AUC of the outlier detection with CAE$\_$KNN, while the solid red line shows the mean AUC of the outlier detection with attCAE$\_$KNN. The lighter shadings present the standard deviation of the three results from the three training processes. One can see that the AUC of CAE$\_$KNN and attCAE$\_$KNN reach the maximum values when the feature number of the embedding layer is set to 20, which is, therefore, chosen to be the number of latent features in CAE and attCAE. In addition, it can also be seen that the stability of attCAE$\_$KNN is higher than that of CAE$\_$KNN.

The detailed implementation of KNN outlier detection refers to the modules in \citep{yy27+zhao2019pyod, yy28+ramaswamy2000efficient}, where there is a core procedure, namely \textit{computeOutliersIndex}. The output data of procedure \textit{computeOutliersIndex} is stored in a heap structure \citep{lattner2003data}.
We take the top 533 galaxy images with the largest values in a heap as outliers and then evaluate the model's performance based on the 533 outliers.

\subsection{Evaluation Metrics}
Besides AUC, we also employ 
Recall, F1 score, and Accuracy to estimate the performance of outlier detection \citep{yy16+cheng2020identifying, yy26+zhu2019galaxy, yy31+hou2019image, yy32+kamalov2020outlier}
, which are given by Eq. (6), (7),(8) and (9).
\begin{equation}
    \label{eq6}
   precision=\frac{TP}{TP+FP} ,
\end{equation}
\begin{equation}
    \label{eq7}
   recall=\frac{TP}{TP+FN} ,
\end{equation}
\begin{equation}
    \label{eq8}
   f1 = 2\times\frac{precision\times recall}{presion + recall} ,
\end{equation}  
\begin{equation}
    \label{eq9}
   accuracy=\frac{TP+TN}{TP+FP+TN+FN} .
\end{equation}  
Be worth mentioning, though Accuracy and F1 score are two of major performance metrics in many applications, they are considered supplements to AUC and Recall since the data distributions in this study are unbalanced (the ratio of the outliers is only $10\%$). In addition , $TP$+$FN$ is equal to the $TP$+$FP$ in all experiments, resulting in the values of $recall$ being equal to the values of F1.

\subsection{Implementation Details}
\label{sect:4.4}
The experimental environment of this study is as follows. We mainly use an Intel Xeon E5-2690 CPU and an Nvidia Tesla K40 GPU. Software environment includes python 3.5, Keras 2.3.1, NumPy 1.16.2, Matplotlib 3.0.3, scikit$\_$learn 0.19.1, and pyod 0.8.4. It takes less than half an hour to train 17778 images running on two NVIDIA Tesla K40 GPUs. 

When training the CAE and attCAE, we set the batch$\_$size to 128 and set epoch to 100, use the binary$\_$crossentropy as described in Eq. (\ref{eq3}) and the Adam as optimizer. One can refer to the settings of the CBAM in \citep{woo2018cbam}.

\section{RESULTS}
\label{sect:results}
This section presents the results of experiments described in Sect.~\ref{sect:experiment}. The outcomes of each experiment and the comparative analysis are listed in the following two subsections.

\subsection{The Case of \textit{Single Type Inliers And Single Type Outliers}}
\label{sect:5.1}

\begin{flushleft}\textbf{Experiment 1}\end{flushleft}
We apply three methods illustrated in Sect.~\ref{sect:method} to the testing set comprising images of Completely round smooth galaxies as inliers and images of Cigar-shaped smooth galaxies as outliers. Five metrics, i.e., Area under the ROC (AUC), Recall, F1 score, accuracy, and runtime,  are utilized to evaluate outlier detection performance of the three methods. The results are shown in Table~\ref{tab3}. Apparently, the attCAE$\_$KNN approach obtains the best performance in all metrics. For instance, the $recall$  using CAE$\_$KNN is $\sim 31\%$ higher than KNN, reaching 57$\%$, while the $recall$ using attCAE$\_$KNN is $\sim 19\%$ higher than CAE$\_$KNN, which can reach 76$\%$. Notably, the runtime of attCAE$\_$KNN is also superior to other methods, and it is $\sim 4\%$ of that of KNN alone.

\begin{table}[]
\caption{The results of Experiment 1, the bold entries highlight our results.}\label{tab3}
    \centering
\begin{tabular}{llllll}
\hline
                     & AUC           & recall        & f1            & acc           & Time                \\ \hline
KNN                  & 0.83          & 0.26          & 0.26          & 0.85          & \textgreater{}4hour \\ \hline
CAE$\_$KNN             & 0.94          & 0.57          & 0.57          & 0.91          & 10min               \\ \hline
\textbf{attCAE$\_$KNN} & \textbf{0.97} & \textbf{0.76} & \textbf{0.76} & \textbf{0.95} & \textbf{10min}     \\ \hline
\end{tabular}
\end{table}

\begin{flushleft}\textbf{Experiment 2}\end{flushleft}
The second testing set contains images of Completely round smooth galaxies as inliers and images of Edge-on galaxies as outliers, as for the experimental parameters are similar to experiment 1. As is shown in Table~\ref{tab4}, the results are similar to experiment 1 too. One of the reasons is that the differences between inliers and outliers are both well distinguished in the first and second experiments.

\begin{table}[]
\caption{The results of Experiment 2, the bold entries highlight our results.}\label{tab4}
    \centering
\begin{tabular}{llllll}
\hline
\textbf{}            & AUC           & recall        & f1            & acc           & Time                \\ \hline
KNN                  & 0.83          & 0.25          & 0.25          & 0.85          & \textgreater{}4hour \\ \hline
CAE$\_$KNN             & 0.95          & 0.56          & 0.56          & 0.91          & 10min               \\ \hline
\textbf{attCAE$\_$KNN} & \textbf{0.98} & \textbf{0.78} & \textbf{0.78} & \textbf{0.96} & \textbf{10min}     \\ \hline
\end{tabular}
\end{table}

\begin{flushleft}\textbf{Experiment 3}\end{flushleft}
This experiment is similar to the previous one, except we adopt images of In-between smooth galaxies as outliers. However, as is shown in Table~\ref{tab5}, the experimental results are different from previous ones because the similarity between inliers and outliers in this test set is less significant than the previous ones. Though including CAE and attention mechanism brings improvement, it is less considerable than the first two cases. For example, concerning $recall$, the CAE$\_$KNN is $\sim 4\%$ higher than KNN and only reaches 15$\%$, while the $recall$ of using attCAE$\_$KNN is higher than CAE$\_$KNN by $\sim 7\%$ and reaches 22$\%$ only. It reveals that the definition of the outliers detection problem is crucial for outlier detection. According to the results, identifying smooth elliptical galaxies with specific ellipticity is not a practical outlier detection problem.

\begin{table}[]
\caption{The results of Experiment 3, the bold entries highlight our results.}\label{tab5}
    \centering
\begin{tabular}{llllll}
\hline
                     & AUC           & recall        & f1            & acc           & Time                \\ \hline
KNN                  & 0.64          & 0.11          & 0.11          & 0.82          & \textgreater{}4hour \\ \hline
CAE$\_$KNN             & 0.68          & 0.15          & 0.15          & 0.83          & 10min               \\ \hline
\textbf{attCAE$\_$KNN} & \textbf{0.71} & \textbf{0.22} & \textbf{0.22} & \textbf{0.84} & \textbf{10min}     \\ \hline
\end{tabular}
\end{table}
\begin{flushleft}\textbf{Experiment 4}\end{flushleft}
Similarly, we adopt images of Spiral galaxies as outliers in this experiment. As expected (see Table~\ref{tab6}), this experiment's results are better than experiment 3 but worse than experiments 1 and 2 because the distinguishability between inliers and outliers in this testing set is more noticeable than that in case 3 but less than cases 1 and 2 (mainly due to the PSF smearing). The most noteworthy difference between completely round-smooth and face-on Spiral galaxies is detailed structures and colors; thus, we hope the improvement of CAE and attention mechanism would be more significant when applying the methods to data from space-born telescopes.

\begin{table}[]
\caption{The results of Experiment 4, the bold entries highlight our results.}\label{tab6}
    \centering
\begin{tabular}{llllll}
\hline
                     & AUC           & recall        & f1            & acc           & Time                \\ \hline
KNN                  & 0.68          & 0.15          & 0.15          & 0.83          & \textgreater{}4hour \\ \hline
CAE$\_$KNN             & 0.77          & 0.24          & 0.24          & 0.84          & 10min               \\ \hline
\textbf{attCAE$\_$KNN} & \textbf{0.81} & \textbf{0.29} & \textbf{0.29} & \textbf{0.86} & \textbf{10min}     \\ \hline
\end{tabular}
\end{table}

\subsection{The Case of \textit{Single Type Inliers And Multiple Type Outliers}}
\label{sect:5.2}

The above experiments primarily explore the feasibility of unsupervised approaches for outlier detection with testing sets containing single type inliers and single type outliers. This sub-section demonstrates an experiment in a more realistic case, i.e., the testing set contains a single type of inliers plus multiple types of outliers. 

\begin{flushleft}\textbf{Experiment 5}\end{flushleft}
We consider images of Completely round smooth galaxies as inliers and images of In-between smooth, Cigar-shaped smooth, Edge-on and Spiral as outliers. The experimental results are shown in Table~\ref{tab7perc}, the attCAE$\_$KNN still achieves the best performance. The $recall$ of CAE$\_k$NN reaches 43$\%$, $\sim 21\%$ higher than KNN, and the $recall$ of attCAE$\_$KNN is $\sim 10\%$ higher than CAE$\_$KNN, reaching to 53$\%$. It is easy to conclude that the missing points in this experiment are dominated by In-between smooth galaxies.

Notably, recall and f1 values are the same in all the experiments since we define the most distant $10\%$ objects to the center of the cluster of inliers in feature space as outliers during the detection of outliers, while the fraction of outliers in the testing set is $10\%$. Consequently, FN equals FP, then $recall$ will equal $precision$, and hence $recall$ equals $f1$ as well. However, when the chosen fraction does not equal the actual value, $recall$ and $f1$ are not the same. The actual fraction of outliers is unknown in real cases; thus, it is impossible to choose a perfect fraction, and one needs to choose a rational fraction to define outliers according to specific scientific goals. We set the fraction to be $5\%$ and $15\%$, in addition to illustrating comparative results. As is shown in Table~\ref{tab7perc}, when the definition of outliers is the most distant $5\%$ objects to the center of the inlier cluster, $recall$ decreases to 0.37, $precision$ increases to 0.74, and $f1$ is 0.5. Whereas, when the definition of outliers is the most distant $15\%$ objects to the center of the inlier cluster, $recall$, $precision$, and $f1$ become 0.67, 0.44, and 0.53 separately. Accordingly, if one plans to obtain a sample of outliers with high completeness, a greater fraction (e.g., $15\%$) is needed, while if the goal is to find rare objects with noticeable and wired features efficiently, a lower fraction (e.g., $5\%$) is practical.

\begin{table}[]
\caption{The results of Experiment 5, the bold entries highlight our results.}\label{tab7perc}
    \centering
\begin{tabular}{lllllll}
\hline
                     & AUC           & recall           & precision             & f1            & acc           & Time \\ \hline
KNN                  & 0.77          & 0.22             & 0.22                  & 0.22          & 0.84          & \textgreater{}4hour \\ \hline
CAE$\_$KNN           & 0.85          & 0.43             & 0.43                  & 0.43          & 0.87          & 10min               \\ \hline
\textbf{attCAE$\_$KNN 5$\%$} & \textbf{0.87} & \textbf{0.37} & \textbf{0.74} & \textbf{0.50} & \textbf{0.92} & \textbf{10min}     \\ \hline
\textbf{attCAE$\_$KNN 10$\%$} & \textbf{0.87} & \textbf{0.53} & \textbf{0.53}  & \textbf{0.53} & \textbf{0.92} & \textbf{10min}     \\ \hline
\textbf{attCAE$\_$KNN 15$\%$} & \textbf{0.87} & \textbf{0.67} & \textbf{0.44} & \textbf{0.53} & \textbf{0.88} & \textbf{10min}     \\ \hline
\end{tabular}
\end{table}

\section{DISCUSSION AND CONCLUSIONS}
\label{sect:discussion}
In this study, we explore the feasibility of applying unsupervised learning to detect outliers in the data of galaxy images. Firstly, we construct three methods, which are built upon the KNN, CAE $+$ KNN, and attCAE$\_$KNN separately. To evaluate the performance of the approaches, we organize two sorts of datasets based on the data of galaxy images given by the project of galaxy zoo challenge published on Kaggle. One group of testing sets includes inliers containing a single type of galaxy morphology plus outliers containing a single type of galaxy morphology; the other group of testing sets is similar to the first ones for inliers but with multiple types of galaxy morphology in the outliers. Comparing the results of applying three approaches to all the testing sets, we find that attCAE$\_$KNN achieves the best performance and costs the least runtime, though its superiority is limited in the case of the testing set with a substantial similarity between inliers and outliers.

Specifically, KNN is usable for outlier detection, but its performance and efficiency are deficient. For instance, the best recall is 0.25, even when the testing set (testing set 1) has significant differences between inliers and outliers. The main reason for the shortcomings is the outdated procedure for extracting features. Therefore, we involve CAE as a module for feature extractions, and then the recall reaches 0.56 in the case of testing set 1. We further employ the attention mechanism to improve the stability of the feature extraction module, and the best recall goes to 0.78 in the case of testing set 1. Repeating the above process in other testing sets that contain single type inliers and single outliers, attCAE$\_$KNN performs the best and costs the least runtime, and one can see more details in Table~\ref{tab3}, Table~\ref{tab4}, Table~\ref{tab5}, Table~\ref{tab6}. 

To test the feasibility of the three methods in a more realistic context, we create testing set 5, containing single type inliers and multiple types outliers. As is expected, attCAE$\_$KNN is still superior to the other two methods. For instance, its recall is 0.53, but the recalls of CAE$\_$KNN and KNN are 0.43 and 0.22, respectively. As is shown in Table~\ref{tab7perc},  the advantage of attCAE$\_$KNN is evident over all five metrics. Hence, we can conclude that outlier detection in galaxy images is feasible by combing CAE and KNN, and the performance can be enhanced by involving the attention mechanism further. Besides, we implement a comparative investigation with different definitions of outliers when detecting them with our methods. The results in Table~\ref{tab7perc} demonstrate that a tighter definition of outliers leads to higher precision but lower recall, while a looser definition of outliers leads to lower precision but higher recall; nevertheless, the overall AUC is stable.

The structures of the testing sets used in the paper are relatively simple compared to real observations since we focus on assessing the feasibility of unsupervised approaches. To make our unsupervised approach suitable for real observations, we are forming a module to reduce any complex case (multiple types inliers $+$ multiple types outliers) to the simple one employed in this paper (single type inliers $+$ multiple types outliers) by combining human inspection and supervised learning. Then, we will apply the pipeline to actual survey data, such as KiDs, DES, and DESI legacy imaging surveys, to test its applicability and reliability. Also, to further improve the performance of approaches, particularly attCAE$\_$kNN, we plan to optimize the architectures and hyper-parameters while applying them to observational data. Last but not least, defining the boundary of inliers and outliers is key to the outlier detection task, as is shown in the results in testing set 3. Hence,  we will adopt a data-driven strategy to investigate the optimal definition of the boundaries according to specific scientific purposes. 

In summary, unsupervised approaches, especially when we involve CAE and the Attention mechanism, are feasible for outlier detection in the datasets of galaxy images. It is foreseen that unsupervised approaches can mine astronomical outliers so as to expand the boundary of human knowledge of the Universe in the big data era. On the other hand, the unsupervised approaches can also detect misclassified samples in standard supervised classification, similar to outlier detection, with no additional efforts. Accordingly, an ideal pipeline for classifying astronomical objects might need to combine supervised and unsupervised manners.

\normalem
\begin{acknowledgements}
The dataset used in this work is collected from the Galaxy-Zoo-Challenge-Project posted on the Kaggle platform. We acknowledge the science research grants from the China Manned Space Project with NO.CMS-CSST-2021-A01 and NO.CMS-CSST-2021-B05. YH, ZQZ, and YLC are thankful for the funding and technical support from the Jiangsu Key Laboratory of Big Data Security and Intelligent Processing. 

\end{acknowledgements}
  
\bibliographystyle{raa}
\bibliography{bibtex}

\end{document}